# DIGITAL RESTORATION OF ANCIENT PAPYRI

A. Sparavigna
Dipartimento di Fisica, Politecnico di Torino
Corso Duca Degli Abruzzi 24, Torino, Italy

**Abstract**
Image processing can be used for digital restoration of ancient papyri, that is, for a restoration performed on their digital images. The digital manipulation allows reducing the background signals and enhancing the readability of texts. In the case of very old and damaged documents, this is fundamental for identification of the patterns of letters. Some examples of restoration, obtained with an image processing which uses edges detection and Fourier filtering, are shown. One of them concerns 7Q5 fragment of the Dead Sea Scrolls.

**Introduction**
In a recent paper [1], we discussed the role of digital restoration in the discovery of a possible Leonardo da Vinci self-portrait. This is a drawing in one of his notebooks, partially hidden by handwriting. The restoration, that removed the text and revealed the drawing, was performed on a digital image of that document, not on the document itself [2,3].
The discovery of a young Leonardo self-portrait, which is of course very important by itself, can be a strong impulse to use image processing for restoration of ancient documents. Actually, the scientific literature is just reporting few applications [4,5]: a reason could be a lack of correlation between the two disciplines, that of image processing and that of ancient manuscript and codices study and preservation.
In [1], we proposed a procedure based on histograms, to remove the writing and reveal the background, as, in the case of Da Vinci's portrait, it was the background that contained the portrait. Let us discuss here another problem, connected with written texts, and in fact opposite to the case of Leonardo portrait. The problem is the reduction of background signals in order to enhance readability of the text. In the case of very old and damaged documents, this is fundamental for identification of the patterns of letters. Here we approach the analysis of old papyri, with image edges enhancing and threshold and Fourier filtering. For the detection of edges we used the dipole moments method [6,7], based on the pixel colour tones distribution. In this present preliminary discussion, we will also propose an elaboration of fragment 7Q5 of the Dead Sea Scrolls.

**The papyrus substrate**
Papyrus was a product firstly manufactured in Egypt. It is relatively cheap and easy to produce, but fragile and sensible to moisture and dryness. Unless the papyrus is of good quality, the writing surface is irregular, and this is an important factor to determine the pattern of letters and compare letters written in different places of the same papyrus [8-10].
We can find volumes in scroll and codices for the following reasons. Papyrus is not strong enough to fold without cracking and a long roll, or scroll, is required to create large volume texts. In the first centuries BC and CE, another material, the parchment, gained importance as a writing surface. Sheets of parchment can be folded to form quires from which codices are fashioned. Writers soon adopted this codex form, and in the Roman world it became common to cut sheets from papyrus rolls in order to form codices. As an improvement on scrolls, the codex form was definitively adopted for texts.
We will consider here examples with papyrus as background.

**Enhancing the letters**

Let us start to work on the image in Fig.1. The image, adapted from Ref.11, shows a piece of a manuscript in Greek on papyrus, from Alexandria, Egypt, 3rd c. BC, written in fine regular Greek uncial. The text is from the poem "Works and Days", by Hesiod. This is by far the earliest surviving manuscript of Hesiod's poem, and also one of the earliest manuscripts of Greek literature. From the image, we can obtain a colour tone map of pixels. The map is then represented by a bidimensional function of pixel position.

At a first glance, it seems possible to use a simple thresholding procedure. Thresholding is the simplest method of image segmentation. From a greyscale image, thresholding can be used to create a binary image: that is, after choosing a specific tone as threshold, higher tones are replace with white and those, which are lower, with black. Applied to Fig.1, this procedure is able to remove the main part of the background, but it removes also some faint letters in the text, therefore reducing the readability of it. We decided then to try to combine thresholding with the detection of edges in the image. The edges have been determined by means of a recently proposed procedure, based on magnitudes of image dipole moments [6,7]. In Fig.1 on the right, it is possible to see the result of edge detection, superimposed to the black letters of text. Mixing this image with the original one, we obtain a more natural result, where the patterns of letters are restored on the papyrus substrate (see Fig.2).

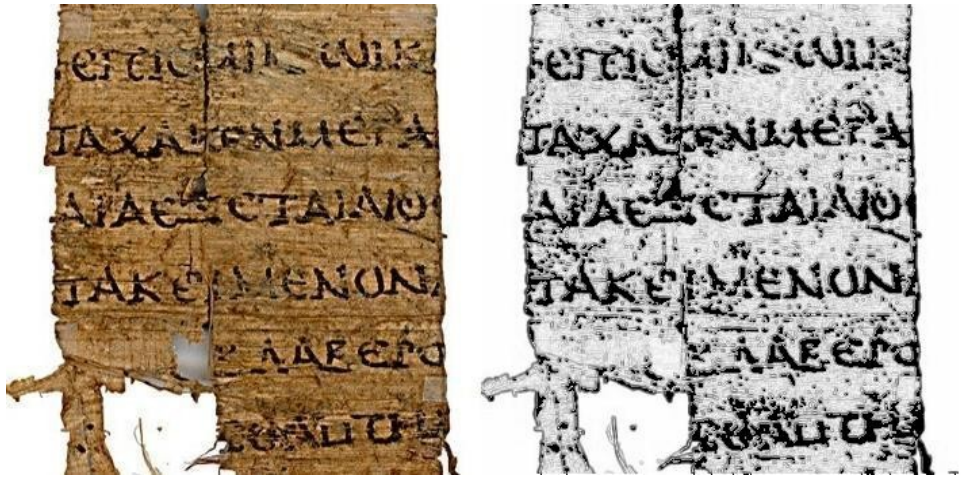

Fig.1 On the left, a manuscript in Greek on papyrus, from Alexandria, 3rd century BC, [11], written in uncial. The edge detection enhances the patterns of letters, as shown by image on the right.

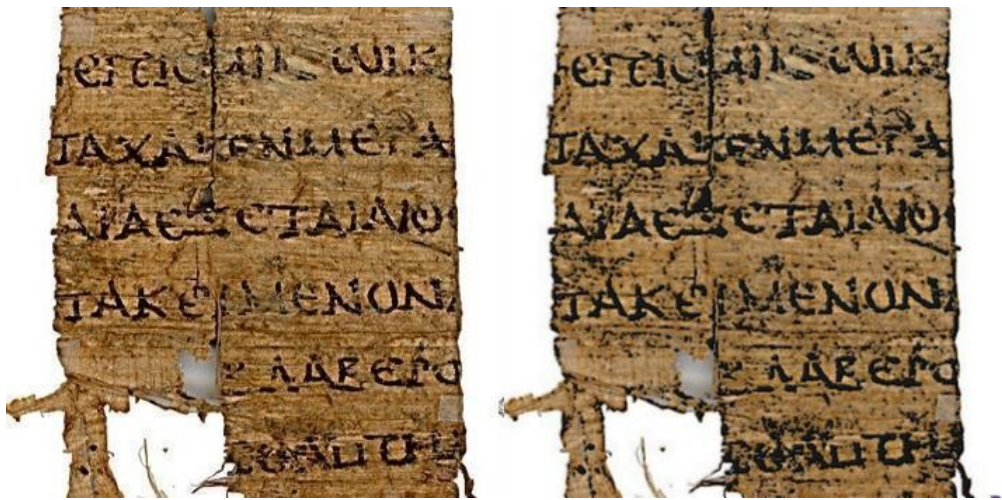

Fig.2 Mixing original image with the result of edge detection give us the image on the right. We obtain a digital restoration of the patterns of letters.

Another example of restoration is proposed in Fig.3: the starting image is a piece of a manuscript in Greek, from Alexandria, Egypt, late 2$^{nd}$ - 1$^{st}$ century BC. The writing is in small Greek uncial. The text is from "Epidemics II", by Hippocrates [11].
Of course, the quality of this proposed digital restoration could be increased by pattern recognition of letters. The images shown in Fig.1 and 2 contain two well-known texts and do not rise questions on text interpretation. But there are circumstances where help in digital enhancing and recognition of the text can be very useful. A case is discussed in the next section.

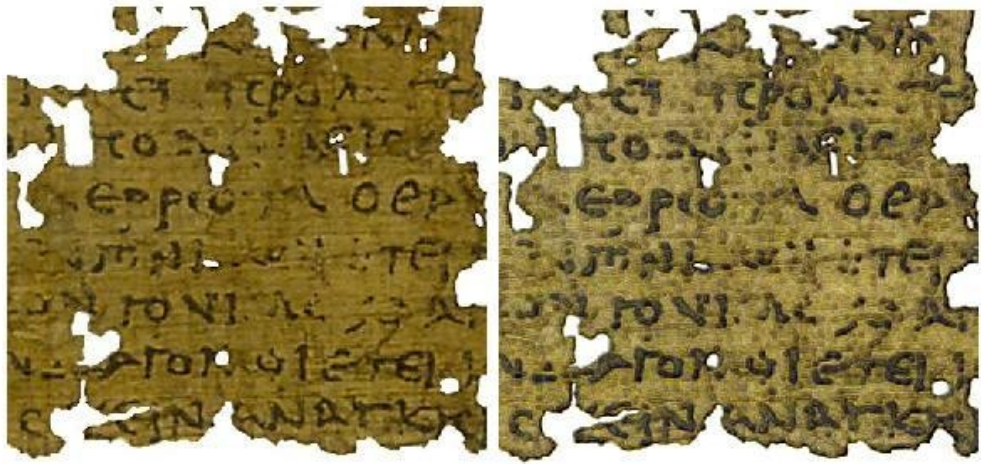

Fig.3 On the left, another manuscript in Greek from Alexandria, Egypt, written in uncial. On the right, a digital restoration obtained using edge detection.

**The 7Q5 fragment**
A collection of scrolls, known as Dead Sea Scrolls, was discovered between 1947 and 1956 in eleven caves around the Wadi Qumran, near the ruins of an ancient settlement on the shore of the Dead Sea. According to a popular theory, the Essenes, a monastic sect, occupied the site. The scrolls, which comprise the oldest known copies of the Hebrew Bible, are widely considered the most important archaeological find of the 20th century. The collection of scrolls includes some of the only known surviving copies of Biblical documents made before 100 CE. They include texts written in Hebrew, Aramaic and Greek, mostly on parchment, some written on papyrus. These manuscripts generally date between 150 BC to 70 CE [12]. Most of the scrolls, probably hidden in the caves during the First Jewish Revolt, are now housed in the Shrine of the Book in Jerusalem. The interest on these manuscripts has been recently revived by new infrared analysis [13].
Among the Dead Sea Scrolls, a small papyrus fragment, named 7Q5, found in the Qumran Cave 7 contains a text in Greek. The significance of this fragment is coming from an assertion of J. O´Callaghan in 1972, appreciated by C.P. Thiede in 1982 [14,15]: the assertion is that 7Q5 is actually a fragment of the Gospel of Mark, 6:52-53. Many scholars have not been convinced by O'Callaghan's and Thiede's identification and rejected this interpretation and proposed other sources for the text [16].
If 7Q5 were identified as Mark 6:52-53 and, as deposited in the cave at Qumran before 68 CE, it would become the earliest known fragment of the New Testament. O'Callaghan's identification is based on the Greek text of Mark 6:52-53. The identification of a letter as a "N" has been strongly disputed because it does not fit into the pattern of another "N" clearly written in another place of the text. This is not a conclusive fact, because there are examples where the pattern of a letter changes in the same text.

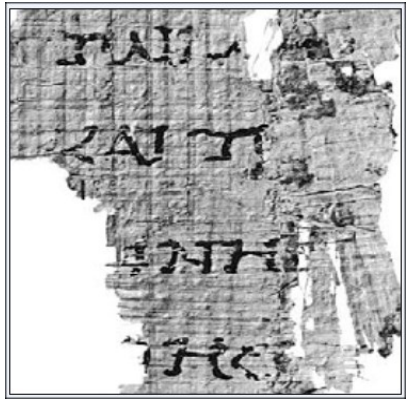

Fig.4 7Q5 is a fragment of the Dead Sea Scrolls (image adapted from [17]).

**The quality of the substrate**
The quality of 7Q5 papyrus is quite different from that of the two previous examples. Horizontal and vertical lines are now clearly visible. Writing on a so irregular surface was not easy for the writer. Of course, this is an important factor to consider in the determination of pattern of letters and in the comparison of letters written in different places of the manuscript. For instance, let us consider the identification of letter "N" in the text. According to O'Callaghan, the letter after "omega" - see the first line of Fig.4 - is "N"; other scholars guess that this letter is "iota". Moreover, the hole after this letter constitutes the problem for a definitive interpretation.
Analysis of Israeli Police showed that there is a faint diagonal line near the top of the left vertical stroke of ink (see the microscope image in Fig.5, adapted from Ref.18). We could imagine, as the substrate surface is not smooth, that our writer had a problem with pen and ink during the writing.
If this faint line were an ink trace, it is possible to guess that during the writing, the "N" diagonal turned out to be broken on such an irregular substrate. To verify this hypothesis, let us apply a high-pass Fourier filtering to the microscope image. The result of filtering is shown in Fig.5 on the right: the faint trace seems to be under the ink, then belonging to a defect of the substrate, which continues under the ink pattern. This fact was previously reported in Ref.18, where the author deduced the letter was a "iota". For filtering we used a FFT program from Ref.19.

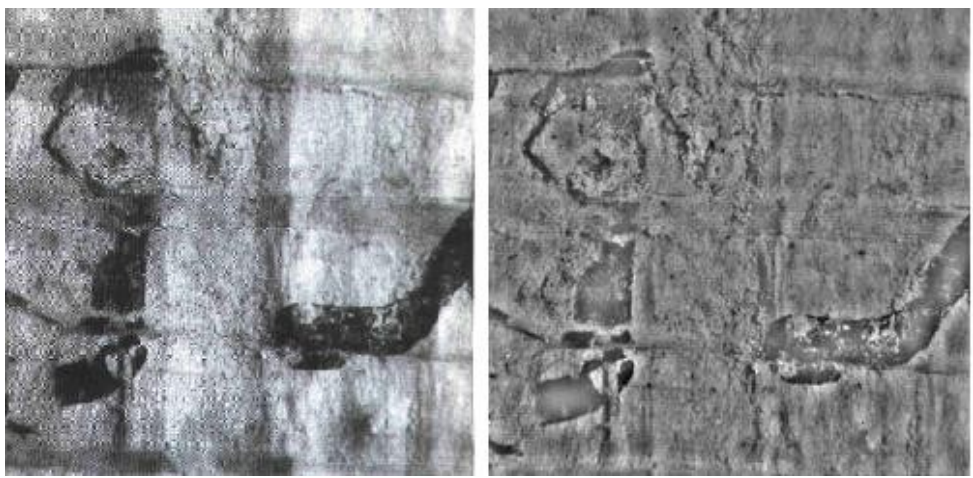

Fig.5 According to O'Callaghan, the letter is a "N"; other scholars guess that this letter is a "iota". Near the top of the left ink stroke, there is a faint diagonal line, which seems to be an ink trace. On the right, the result obtained by applying a high-pass Fourier filter. The faint line seems to be under the ink pattern of the letter, and then to be a structure of the substrate.

The actual identification of letters in 7Q5 is beyond the scope of this paper. We have considered 7Q5 to show how digital image processing can be useful in discriminate ink and substrate.
Let us come back to a digital restoring of 7Q5. We can try to remove the background texture. A custom Fourier filtering of image in Fig.4 - by means of a FFT program [19] - gives the image in Fig.6, on the right. This filtering removes papyrus vertical lines.

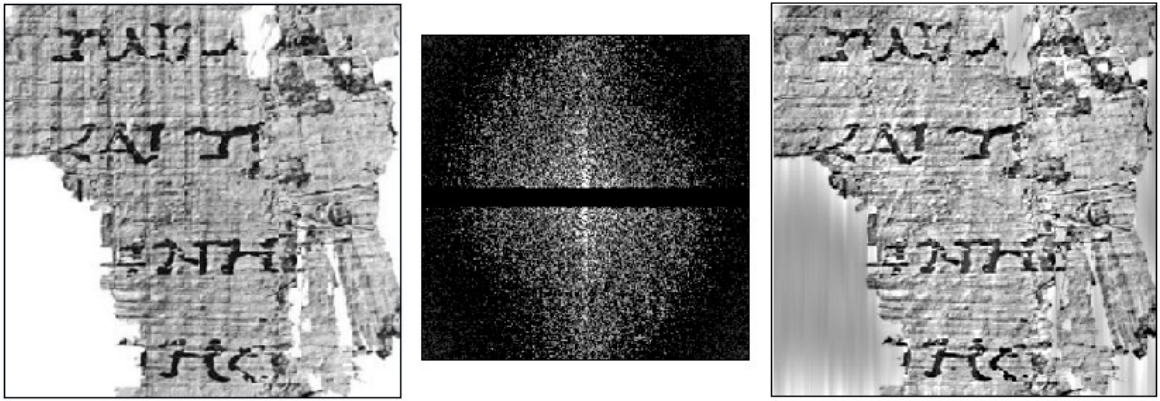

Fig.6 On the left, the original image is shown. In the middle, the custom Fourier filtering used to obtain the image on the right: this filtering removes vertical lines of papyrus.

Digital restoration can be performed on the filtered image. Again, image edges are detected and the map of edges superimposed on starting image. The final result is shown in Fig.7, where we can see a rendering of the image as a bas-relief.
To remove the background texture, it would be better to work on a colour figure, if it is at disposal for processing, because background has usually colour tones different from those of ink. For 7Q5, we have not a good colour image, unfortunately.

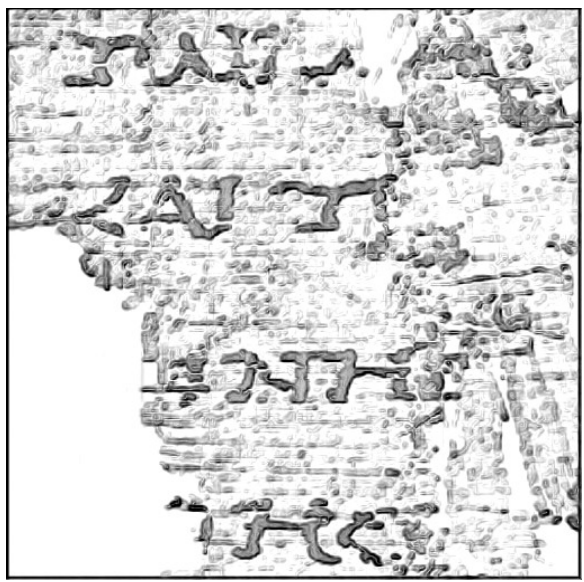

Fig.7 A proposed rendering of 7Q5.

## Conclusions

We started a discussion on the use of image processing in restoration of old documents. This kind of restoration, which is not made on the document itself but on its digital image, can give excellent results and be very useful in pattern recognition of hand-written letters. We proposed a procedure based on edge detection, a procedure that can be further increased in quality.

We used also a Fourier filtering, for a preliminary filtering of images to remove the periodic structures of the background. On filtered images, the digital restoration can help in enhancing the text visibility and interpretation. Applied to a microscope image of 7Q5 fragment, the filtering seems to provide new evidences to evaluate the patterns of ink and substrate.

## Acknowledgement

The author would like to thank Professor Marco Omini for his interest in this work and his constant encouragement. The author is indebted to him for his suggestion to investigate the Dead Sea Scrolls and Greek manuscripts.